\pdfoutput=1 
\documentclass{article} % For LaTeX2e
\usepackage{gmas_iclr2022_conference,times}

\usepackage{amsmath,amsfonts,bm}

\def\eqref#1{equation~\ref{#1}}
\def\1{\bm{1}}

\DeclareMathAlphabet{\mathsfit}{\encodingdefault}{\sfdefault}{m}{sl}
\SetMathAlphabet{\mathsfit}{bold}{\encodingdefault}{\sfdefault}{bx}{n}

\usepackage{hyperref}
\usepackage{url}
\usepackage{subcaption}
\usepackage{graphicx}

\title{Deep Learning Agents Trained for Avoidance Behave Like Hawks and Doves}

\author{Aryaman Reddi\\
Department of Engineering\\
University of Cambridge\\
Cambridge, United Kingdom\\
\texttt{adr43@cantab.ac.uk,aryamanreddi@gmail.com} \\
}

\iclrfinalcopy % Uncomment for camera-ready version, but NOT for submission.
\begin{document}

\maketitle

\begin{abstract}
 Effective navigation and obstacle avoidance are desired features of successful multi-agent systems. Inter-agent avoidance presents additional challenges due to the dynamic nature of such tasks, as well as the non-stationary behavior of online learning agents. In this work, we study the behaviors learned by two reinforcement learning agents in a grid world that must cross paths to reach a goal destination without interfering with each other. Our findings indicate that agents reach an asymmetric routing protocol that exhibits similar payoffs to the `Hawks and Doves' game, in that one agent learns an aggressive strategy while the other learns to work around the aggressive agent. 

% We present heuristically optimal strategies expressed by deep learning agents playing a simple avoidance game. We analyze the learning and behavior of two agents within a symmetrical grid world that must cross paths to reach a target destination without crashing into each other or straying off of the grid world in the wrong direction. The agent policy is determined by one neural network that is employed in both agents. Our findings indicate that the fully trained network exhibits behavior similar to that of the game Hawks and Doves, in that one agent employs an aggressive strategy to reach the target while the other learns how to avoid the aggressive agent.
\end{abstract}

\section{Introduction}
Multi-agent systems typically involve multiple autonomous agents interacting within a shared environment. These systems show great promise for applications in a wide range of real-world problems, including autonomous vehicles~\cite{zhang2024multi}, robotic swarms~\cite{debie2023swarm}, and home robots~\cite{kim2024armor}. A key challenge in such systems is inter-agent navigation - ensuring that agents can move within the environment efficiently while avoiding collisions and minimizing disruptions to others. Effective solutions to this problem are essential for improving coordination and safety in dynamic environments.

Reinforcement learning (RL) has produced impressive results in problems such as Chess~\cite{silver2017mastering}, robot learning~\cite{bohlinger2024one}, and network optimization~\cite{jamil2022reinforcement}. RL provides a promising approach for addressing multi-agent navigation by allowing agents to learn optimal avoidance behaviors by optimizing long-term rewards. Multi-agent RL introduces additional challenges, such as non-stationarity due to simultaneously learning agents~\cite{papoudakis2019dealing}.

In this work, we investigate the avoidance protocol used by two trained RL agents in symmetric navigation problems. While inter-agent avoidance problems between humans often utilize systems (e.g. traffic systems and pedestrian walkways) and social protocols (e.g. `always overtake on the right') defined at several levels , these protocols sometimes fail, resulting in hazardous collisions. We find that RL agents trained to solve a symmetric navigation game reach an asymmetric avoidance protocol that achieves a payoff structure similar to the game `Hawks and Doves'~\cite{smith1973logic}, in which one agent always chooses an aggressive strategy while the other agent learns how to work around the aggressive agent. 

\section{Related works}
\label{related}
Imbuing agents with safe and effective navigation policies has remained a key area of study in multi-agent systems research.  \cite{ma2023multi} investigates multi-agent navigation in unconstrained environments using a graph neural network-based controller and a centralized attention mechanism. \citet{li2024multi} proposes a method for dynamic multi-robot navigation using a relational reasoning and trajectory decoder, inferring relations between agent movements and agent cost functions. In a similar vein,~\cite{aljassani2023enhanced} formulates EMAFOA, an obstacle avoidance algorithm that can be used for second-order and non-holonomic multi-agent systems.

RL has been used for producing effective movement policies across several works. \citet{yuan2023survey} provides a comprehensive study of cooperative MARL systems in open environments, including autonomous driving and swarm control. \citet{pinto2017robust, pan2019risk, kamalaruban2020robust, cai2018curriculum, reddi2023robust} enhance agent navigation and locomotion policies using a two-agent system which introduces adversarial disturbances. Additionally,~\cite{mulgaonkar2017robust, roy2016study, hedjar2014real} investigate dynamic obstacle avoidance in robot swarms.

\section{Background}
We formulate a two-agent version of a Markov Decision Process (MDP)~\cite{puterman2014markov} known as a Markov game~\cite{littman1994markov}, defined by a tuple $\mathcal{G}=\langle\mathcal{I},\mathcal{S},\mathcal{A},\mathcal{P},\mathcal{R},\gamma,N,\iota\rangle$. $\mathcal{I}\equiv\{1,2\}$ is the set of agents, $\mathcal{S}$ is the state space, $\mathcal{A} \equiv \times_{i\in \mathcal{I}}A_{i}$ is the joint action space of the agents. At each timestep, agent $i$ uses a policy $\pi_{i}\left(a_{i}|s\right)$ parameterized by $\boldsymbol{\theta}_{i} \in \mathbb{R}^{d_{i}}$ to select an action $\boldsymbol{a}_{i}\in A_{i}$. The joint action of all agents $\boldsymbol{a}$ determines the next state according to the joint state transition function $\mathcal{P}\left( s'|s,\boldsymbol{a} \right)$. $\mathcal{R}\equiv\{R_{1},R_{2}\}$ are the set of agent reward functions. Each agent $i$ has a learning rate $\eta_{i}$ and receives a reward $r_{t,i}$ at time $t$ according to its reward function $R_{i}\left(s_{t}, \boldsymbol{a}_{t}, s_{t+1} \right)$. $\gamma$ is the discount factor and $\iota$ is the initial state distribution.

Each agent aims to maximize its own discounted objective

\begin{equation}
    J_{i}\left(\boldsymbol{\theta}_{1},\boldsymbol{\theta}_{2}\right) = \underset{\tau\sim\ \iota, \pi_{1},\pi_{2},\mathcal{P}}{\mathbb{E}}\left[\sum_{t=0}^{\infty} \gamma^t r_{t,i}\right],
    \label{E:ma_objective}
\end{equation}

where $\tau$ refers to the trajectories induced under the agent policies, $\iota$, and $\mathcal{P}$. The joint stochastic policy $\boldsymbol{\pi}$ induces a joint \textit{action-value function} for each agent

\begin{equation}
    Q_{i}\left(s, \boldsymbol{a}_{1},\boldsymbol{a}_{2}\right) = \underset{\pi_{1}, \pi_{2}}{\mathbb{E}}\left[\sum_{l=t}^{\infty} \gamma^{l-t} r_{l,i} \Big| s, \boldsymbol{a}_{1},\boldsymbol{a}_{2}\right].
    \label{E:q_function}
\end{equation}

For any finite MDP, it can be shown that Q-learning~\cite{watkins1989learning} will converge to the optimal policy given infinite time and sufficient exploration. 

\section{Implementation}
\label{gen_inst}

The game is implemented as a 64x64 grid world. Agents are spawned on the edges of the grid world and must navigate to the opposite side to that where they are spawned.

Each agent is provided a sparse reward for reaching the target edge, while a large penalty is provided for colliding with the other agent or for crossing the grid world boundary on an incorrect side. Additionally, a small penalty for each frame the agent has not reached the target encourages it to seek out shorter paths over time. Each agent perceives the grid world as a 2D image with separate frames for its own position, the position of the other agent, and the boundary which it must reach to win. Each of these frames is implemented with frame stacking to incorporate temporal information, as in~\citet{mnih2015human}. 

\begin{figure}
\begin{center}
\includegraphics[width=0.6\columnwidth]{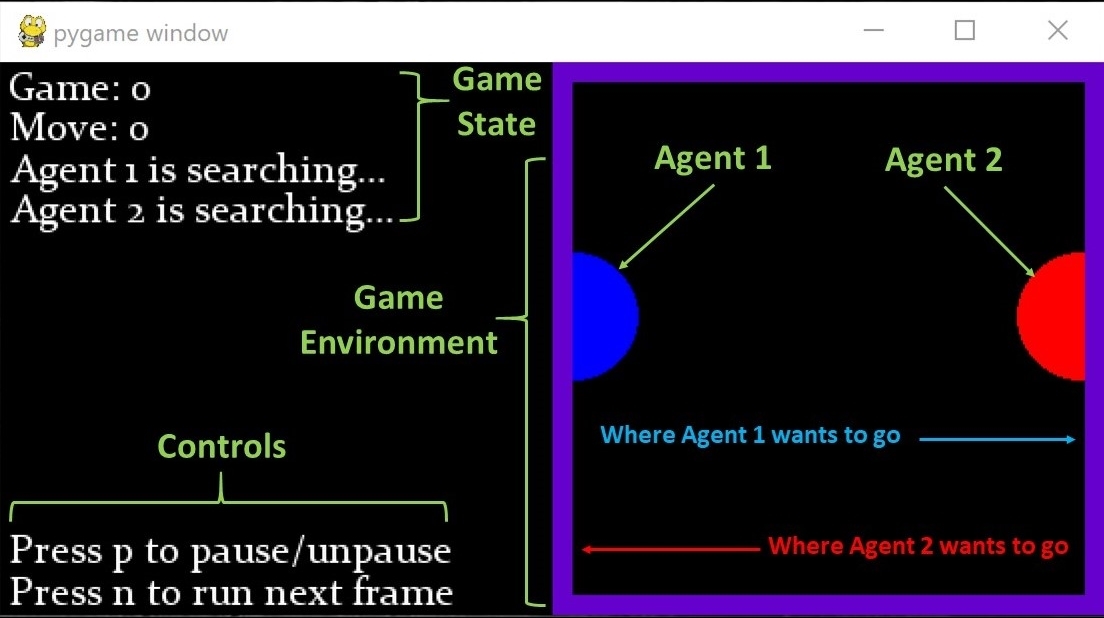}
\end{center}
\caption{Labeled game GUI}
\label{fig:gui}
\end{figure}

\begin{figure}
    \centering
    \subcaptionbox{Parallel Pathways\label{fig:5-para_dia}}
    {\includegraphics[width=4cm]{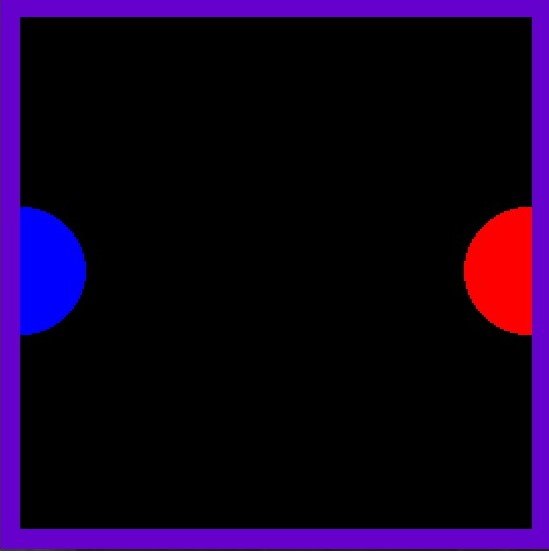}}
    \subcaptionbox{Perpendicular Pathways\label{fig:5-perp_dia}}
    {\includegraphics[width=4cm]{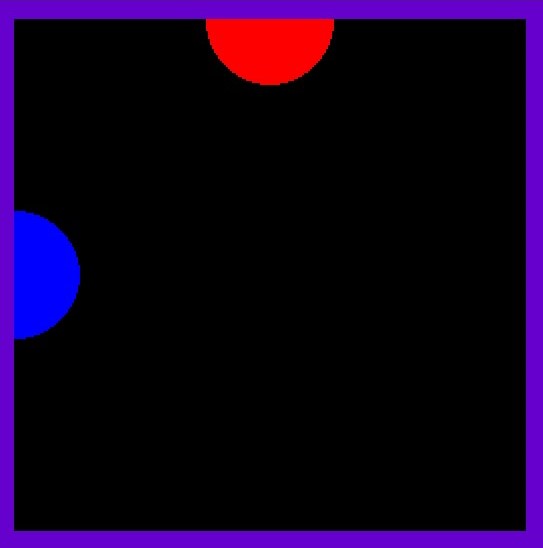}}
    \caption{Multi-agent navigation problem settings}\label{fig:5-ped_configs}
\end{figure}

Figure~\ref{fig:gui} presents the GUI for this game while Figure~\ref{fig:5-ped_configs} presents the two navigation challenges solved by the agents with symmetric starting positions.\footnote{The code to reproduce this project can be found at \url{https://github.com/AryamanReddi99/IIB-Nim-Pedestrians}.}

The agents are implemented using deep Q networks~\cite{mnih2013playing} equipped with convolutional layers for feature extraction. Additionally, the techniques of experience replay and a target network were used to stabilize training~\cite{schaul2015prioritized,zhang2021breaking}.

We are concerned with the two training configurations shown in Figure~\ref{fig:5-ped_configs}. Figure~\ref{fig:5-para_dia} shows the game configuration where the agents face each other and must cross parallel paths, while Figure \ref{fig:5-perp_dia} shows the game configuration where agents must cross perpendicular paths.

\section{Results}
\begin{figure}
    \centering
    \subcaptionbox{Parallel strategies\label{fig:5-para_sol}}
    {\includegraphics[width=4cm]{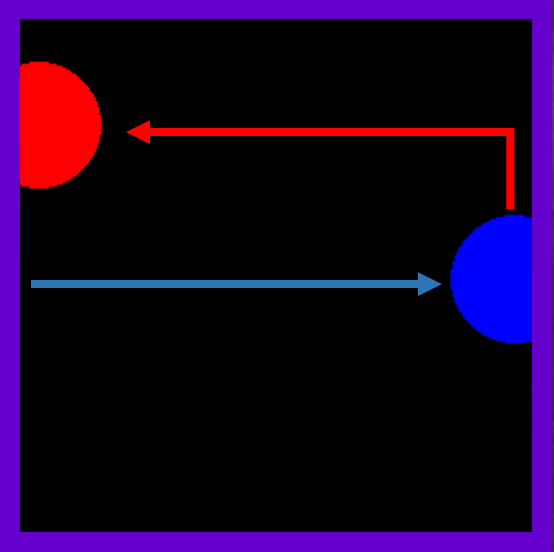}}
    \subcaptionbox{Perpendicular strategies\label{fig:5-perp_sol}}
    {\includegraphics[width=4cm]{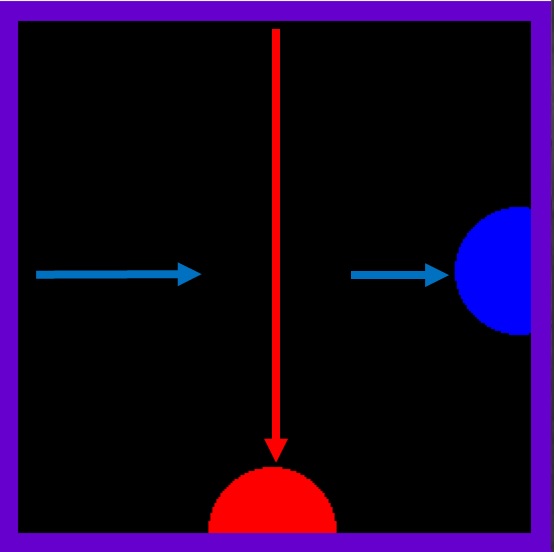}}
    \caption{Converged strategies.}\label{fig:sols}
\end{figure}

It was found that the agents converge on asymmetric strategies when trained with deep Q-learning, despite the symmetry of the configurations in Figure \ref{fig:5-ped_configs}. Figure \ref{fig:sols} illustrates the solutions that the agents mutually converged upon. In practice, the agents initially discover a rough pathway to their targets, and then refine their strategies at training continues. This results in one agent (whichever one finds the sparse reward first) learning to minimize penalties by moving to its target in a straight line, resulting in the highest possible single-episode reward. Meanwhile, the other agent learns to avoid the first agent. 

Figure~\ref{fig:5-para_sol} illustrates this asymmetric protocol for the parallel pathways case: one agent learns to move straight to the opposite side, while the other must step out of the way first to avoid a collision. In the situation in Figure~\ref{fig:5-perp_sol}, we see once again that one agent learns to move straight to the target. Meanwhile, the other agent learns to wait for the first agent to proceed before crossing to its target. Since the agent positions are symmetric, each agent learns its strategy arbitrarily depending on which one finds the sparse reward first.

This result is initially surprising because a mutually energy-efficient symmetric solution is expected given the symmetry of the game - this would manifest in the situation whereby both agents incur equal penalty in avoiding collision. An analysis of the rewards incurred in this game sheds some light on why this mixed strategy profile is more likely to manifest.

\begin{figure}
\begin{center}
\includegraphics[width=0.4\columnwidth]{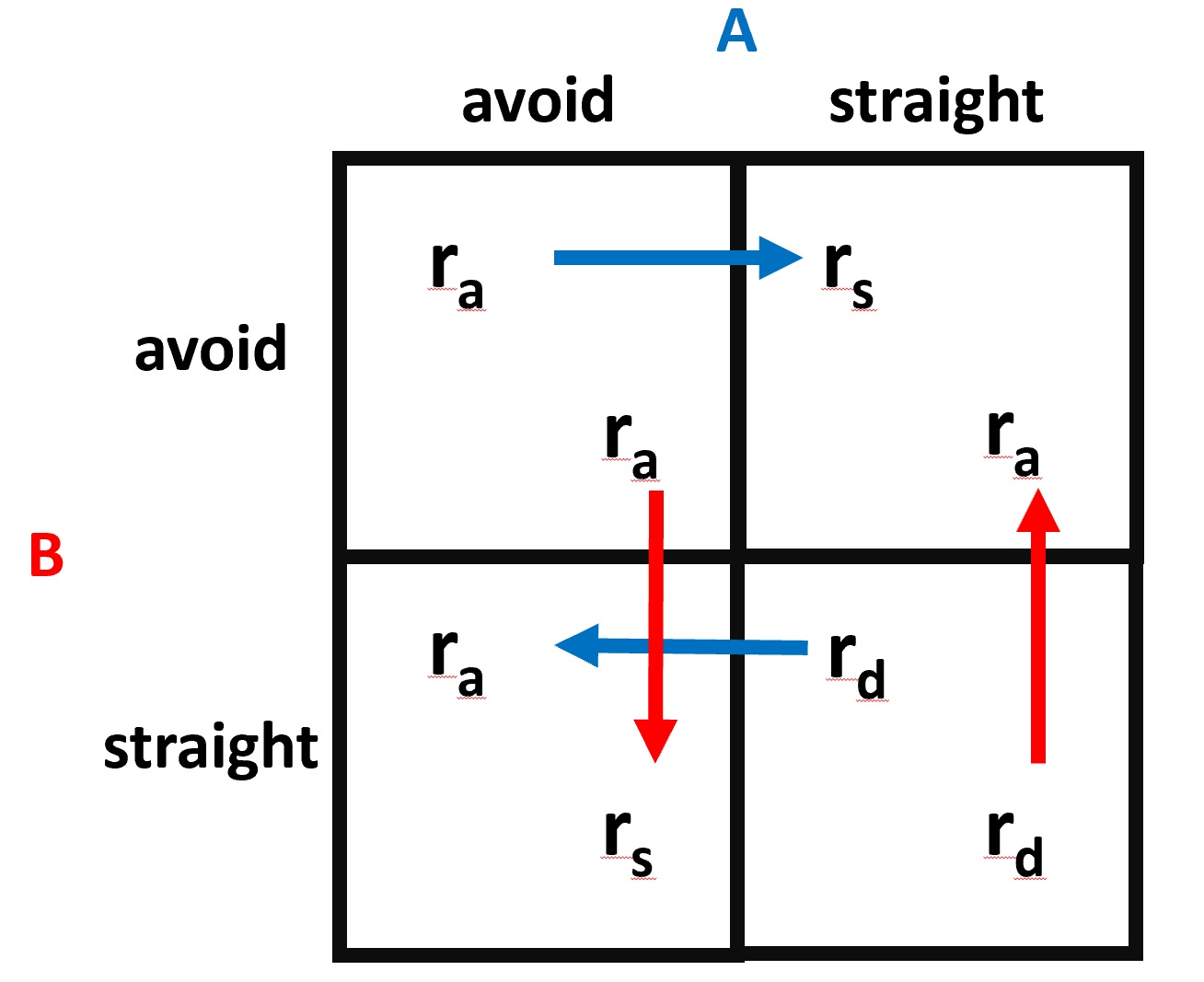}
\end{center}
\caption{Payoff matrix of strategies.}
\label{fig:nash}
\end{figure}

Consider Figure \ref{fig:nash}, which shows the payoff matrix between two agents A and B. Each agent can play either the 'straight' strategy, which takes the shortest possible path to the target, or the 'avoid' strategy, which finds a path that avoids the other agent. The rewards obtained by the agents are shown in each quadrant, with the value in the upper-left corner indicating the reward for agent A and the value in the lower-right corner indicating the reward for agent B. 

When both agents play the 'straight' strategy, they will collide and terminate the episode, resulting in a large negative reward - we denote this $\mathnormal{r_{d}}$ (the reward for death). When an agent reaches its target in a straight line, it will achieve the highest possible reward for a single episode, denoted by $\mathnormal{r_{s}}$. When an agent plays 'avoid', it must spend some time avoiding the other agent instead of moving towards its target, resulting in a reward $\mathnormal{r_{a}}$. Equation \ref{eq:q7} describes the monotonic relationship between the rewards,

\begin{equation} \label{eq:q7}
\mathnormal{r_{s}} > \mathnormal{r_{a}} > \mathnormal{r_{d}}.
\end{equation}

Figure \ref{fig:nash} indicates action preferences for each agent; for each action profile of each agent, an arrow points in the direction of increasing reward for the other agent given that the action profile of the first is fixed. The arrows signify that when the agents play opposing strategies, neither can increase their reward using a unilateral change in strategy, also known as a Nash equilibrium. This is analogous to the equilibrium situation seen in the game Hawks and Doves, which also models a game played between 'aggressive' and 'friendly' strategies~\cite{smith1973logic,grafen1979hawk}.

\section{Conclusion}
In this work we implement a simple two-agent collision avoidance problem and train deep Q-networks to converge on viable strategies. We find that despite the symmetric starting positions of the scenarios, agents do not converge on cost-symmetric policies. Instead, the strategy profile of the agents converges on an asymmetric protocol reminiscent of Maynard Smith's famous `Hawks and Doves' game. This results in one agent always choosing to play the aggressive, high-reward strategy of moving in a straight line to its target, while the other agent plays a cautious strategy which avoids the first agent and incurs a small penalty doing so.

\section{Reproducibility}
The code to reproduce this project can be found at \url{https://github.com/AryamanReddi99/IIB-Nim-Pedestrians}.

\bibliography{gmas_iclr2022_conference}
\bibliographystyle{gmas_iclr2022_conference}

\end{document}